\documentclass[sigconf,nonacm]{acmart}

\usepackage{amsmath}
\usepackage{enumitem}
\usepackage{algorithm}
\usepackage{algpseudocode}
\usepackage{xcolor}

\newcommand{\param}[1]{{\color{blue}#1}}
\newcommand{\lastobid}{\omega^o_{t}}

\newcommand{\uthresh}{\bar{u}_t}

\newcommand{\estOppmodel}{\widehat{U}_o}
\newcommand{\ohist}{\Omega^o_t}
\newcommand{\realUsermodel}{U_u}

\acmBooktitle{XAIFIN2023: Workshop on Explainable AI in Finance,
 November 27, 2023, NY} 
\begin{document}

\title{Towards Explainable Strategy Templates using NLP Transformers}

\author{Pallavi Bagga}
\affiliation{%
  \institution{Moorfields Eye Hospital NHS Foundation Trust}
  \city{London}
  \country{UK}}
\email{pallavi.bagga@nhs.net}

 \author{Kostas Stathis}
\affiliation{%
  \institution{Royal Holloway, University of London}
  \city{Egham}
  \country{UK}
}\email{kostas.stathis@rhul.ac.uk}

\renewcommand{\shortauthors}{P. Bagga \& K. Stathis}

\begin{abstract}
This paper bridges the gap between mathematical heuristic strategies learned from Deep Reinforcement Learning (DRL) in automated agent negotiation, and comprehensible, natural language explanations. Our aim is to make these strategies more accessible to non-experts.
% representation learned from applying Deep Reinforcement Learning (DRL) to produce heuristic strategies in automated agent negotiation, and comprehensible, natural language explanations that make accessible components of that strategy to non-experts. 
By leveraging traditional Natural Language Processing (NLP) techniques and Large Language Models (LLMs) equipped with Transformers, we outline how parts of DRL strategies composed of parts within strategy templates can be transformed into user-friendly, human-like English narratives. To achieve this, we present a top-level algorithm that involves parsing mathematical expressions of strategy templates, semantically interpreting variables and structures, generating rule-based primary explanations, and utilizing a Generative Pre-trained Transformer (GPT) model to refine and contextualize these explanations. Subsequent customization for varied audiences and meticulous validation processes in an example illustrate the applicability and potential of this approach. 
%This research not only enhances the accessibility and practical utility of strategies construed as strategy templates for a diverse array of stakeholders, ranging from financial analysts and negotiators to non-expert users, but also advances the initiative of explainable artificial intelligence within strategic artificial agents in financial and negotiation scenarios. This encourages transparent, trustworthy, and more widely applicable use of such technologies in real-world decision-making environments.
\end{abstract}

\keywords{Automated Negotiation, Explainable Artificial Intelligence, Strategy Templates, Natural Language Processing, Transformers}

\maketitle

{\bf Reference Format:}\\
Pallavi Bagga and Kostas Stathis. 2023. Towards Explainable Strategy Templates using NLP Transformers. In XAIFIN2023: Workshop on Explainable AI in Finance, November 27, 2023, ACM, New York, USA, 6 pages.\\

\section{Introduction}
Many applications are based on transactions involving multiple parties interacting with each other to exchange or (re)allocate goods in electronic markets. Examples include automated trading in finance~\cite{liu2021finrl}, multi-issue negotiation in logistics~\cite{van2006automating}, and others. In many of these settings, to perform these transactions, parties employ complex strategies and tactics that are often mathematical and thus inaccessible to non-expert users. Motivated by our previous work with agent negotiation strategies~\cite{KAIS17, anegma}, this paper explores how to compose complex strategies from tactics and how to transform tactics %learned using Deep Reinforcement Learning (DRL) 
into simple, understandable English sentences. %in order to make the original complex strategic ideas accessible to everyone, including those that do not posses enough technical expertise in the application domain.

Translating complex information into easily digestible forms isn't new. Existing work has explored the use of Natural Language Processing (NLP) and Machine Learning (ML) to improve the explainability of complex concepts~\cite{raffel2020exploring, liu2019roberta, yang2019xlnet}, acting as a bridge between detailed technical data and clear, accessible information. For example, the work by~\cite{bagga2020learnable, bagga2022deep, bagga2021agent} explored ways into creating strategy tactics from learnable strategy templates that allowed a software agent to bid and accept offers to/from an opponent agent during the multi-issue bilateral negotiation, using an actor-critic-based deep learning architectures. %This process created `strategic templates' as  mathematical expressions that constrained the decision making of the agent, 
While providing a solid base in understanding the mathematical roots of strategy templates, the challenge of making this knowledge accessible to those without a technical background is still a notable gap in existing research. 

Building on our previous work with strategy templates, we focus on templates for heuristic negotiation strategies in different domains, learned to accept or offer bids using a deep reinforcement learning (DRL) framework~\cite{bagga2022deep}. We then use NLP techniques and logical rules to transform strategy tactics into English sentences. Specifically, we combine GPT-4~\cite{openai2023gpt4, brown2020language} for language processing, SymPy~\cite{meurer2017sympy} for symbolic mathematics in Python, and spaCy~\cite{honnibal2015spacy} for advanced NLP, to develop a framework aiming to simplify strategy templates. This paves the way towards AI-enhanced strategies that are transparent, accessible, and beneficial to a wide range of stakeholders in domains such as negotiation and, by analogy of the type of transactions supported, finance. 

\section{The ANESIA Model}
In~\cite{bagga2022deep}, we proposed a DRL model for building an actor-critic architecture to support bilateral negotiation for multiple issues. In particular, in this proposal we introduced ``interpretable'' strategy templates as a mechanism to learn the best combination of acceptance and bidding tactics at any negotiation time.
% as in~\cite{bagga2020learnable}. 
Each strategy template consists of a set of parameterized tactics that are used to decide an optimal action at any time. Thus, we built automated agents for multi-issue negotiations that can adapt to different negotiation domains without the need to be pre-programmed. 

\begin{figure*}
    \centering
    \includegraphics[scale=0.45]{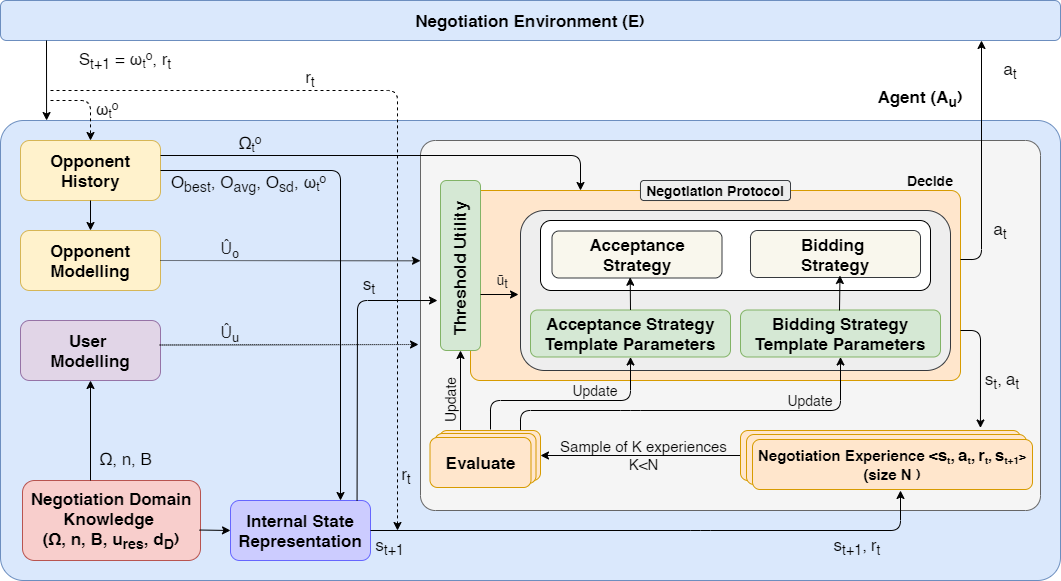}
    \caption{The DLST-ANESIA Agent Negotiation Model}
    \label{anesia-ii}
\end{figure*}

We assumed a negotiation environment $E$ containing two agents $A_u$ and $A_o$ negotiating with each other over some domain $D$ as shown in Figure~\ref{anesia-ii}. A domain $D$ consists of $n$ different independent issues, $D = (I_1, I_2,  \dots I_n)$, with each issue taking a finite set of $k$ possible discrete or continuous values $I_i = (v^i_1, \ldots v^i_{k})$, as in~\cite{baarslag2014decoupling, bagga2021pareto}.

In our experiments, we considered issues with discrete values only. An agent's bid $\omega$ is a mapping from each issue to a chosen value (denoted by $c_i$ for the $i$-th issue), i.e., $\omega = (v^1_{c_1}, \ldots v^n_{c_n})$.  
The set of all possible bids or outcomes is called outcome space $\Omega$, such that $\omega \in \Omega$. The outcome space is common knowledge to the negotiating parties and stays fixed during a single negotiation session.

Before the agents can begin the negotiation and exchange bids, they must agree on a negotiation protocol $P$, which determines the valid moves agents can take at any state of the negotiation ~\cite{fatima2005comparative}. Here, we consider the alternating offers protocol \cite{rubinstein1982perfect}, with possible $Actions = \{offer(\omega), accept, reject\}$.

In this context, we also took as given a collection of acceptance and bidding tactics, $\mathcal{T}_a$ and $\mathcal{T}_b$. Each $\mathtt{t}_a \in \mathcal{T}_a$ maps the agent state, threshold utility, opponent bid history, and a (possibly empty) vector of learnable parameters $\mathbf{p}$ into a utility value: if the agent is using tactic $\mathtt{t}_a$ and $\mathtt{t}_a(s_t, \bar{u}_t, \ohist, \mathbf{p}) = u$, then it will not accept any offer with utility below $u$, see \eqref{eq:acc_templ} below. 
Each $\mathtt{t}_b \in \mathcal{T}_b$ is of the form $\mathtt{t}_b(s_t, \bar{u}_t, \ohist, \mathbf{p}) = \omega$ where $\omega\in \Omega$ is the bid returned by the tactic. 

Given this idea of tactics, an \textit{acceptance strategy template} is a parametric function given by
\vspace*{-\parskip}
\begin{equation}\label{eq:acc_templ}
    \bigwedge_{i=1}^{n_a} t \in [t_i, t_{i+1}) \rightarrow \left( \bigwedge_{j=1}^{n_i}
  { \color{blue}{c_{i,j}}} \rightarrow \widehat{U}(\lastobid) \geq \mathtt{t}_{i,j}(s_t, \bar{u}_t, \ohist, \param{\mathbf{p}_{i,j}})\right)
\end{equation}
where $n_a$ is the number of phases; $t_1=0$, $t_{n_a+1}=1$, and $t_{i+1}=t_i+ \param{\delta_i}$, where the $\param{\delta_i}$ parameter determines the duration of the $i$-th phase; 
for each phase $i$, the strategy template includes $n_i$ tactics to choose from: $\param{c_{i,j}}$ is a Boolean choice parameter determining whether tactic $\mathtt{t}_{i,j}\in \mathcal{T}_a$ should be used during the $i$-th phase. 
Note that~\eqref{eq:acc_templ} is a predicate returning whether or not the opponent bid $\lastobid$ is accepted. 

Similarly, 
a \textit{bidding strategy template} is defined by 
\vspace*{-\parskip}
\begin{equation}\label{eq:bid_templ}
    \bigcup_{i=1}^{n_b} 
    \begin{cases}
    \mathtt{t}_{i,1}(s_t, \bar{u}_t, \ohist, \param{\mathbf{p}_{i,1}}) & \text{ if } t \in [t_i, t_{i+1})  \text{ and } \param{c_{i,1}}\\
    \cdots & \cdots\\
    \mathtt{t}_{i,n_{i}}(s_t, \bar{u}_t, \ohist,  \param{\mathbf{p}_{i,n_i}}) & \text{ if } t \in [t_i, t_{i+1})  \text{ and } \param{c_{i,n}}
    \end{cases}
\end{equation}
where $n_b$ is the number of phases, $n_i$ is the number of options for the $i$-th phase, and $\mathtt{t}_{i,j}\in \mathcal{T}_b$. $t_i$ and $\param{c_{i,j}}$ are defined as in the acceptance template. The particular libraries of tactics used in this work are discussed in the next Section. 
We stress that both \eqref{eq:acc_templ} and~\eqref{eq:bid_templ} describe time-dependent strategies where a given choice of tactics is applied at different phases (denoted by 
$t \in [t_i, t_{i+1})$).

%\textcolor{red}{HERE: The transition between paragraphs here is abrupt. Which negotiation phases do you }

In this work, we are interested in using strategy templates to develop time-dependent, heuristic strategies, which enable the agent to apply a different set of tactics at different time intervals or phases. The number of phases $n$ and the number of tactics $n_i$ to choose from at each phase $i=1,\ldots,n$ are the only parameters fixed in advance. For each phase $i$, the duration $\delta_i$ (i.e., $t_{i+1} = t_{i} + \delta_i$) and the choice of tactic are learnable parameters. The latter is encoded with choice parameters $c_{i,j}$, where $i=1,\ldots,n$ and  $j=1,\ldots,n_i$, such that if $c_{i,j}$ is true then the $(i,j)$-th tactic is selected for phase $i$. Tactics may be in turn parametric, and depend on learnable parameters $\mathbf{p}_{i,j}$. The tactics for acceptance strategies are:
\begin{itemize}
\item $\realUsermodel(\omega_t)$, the estimated utility of the bid $\omega_t$ that our agent would propose at time $t$. 
\item 
$Q_{\realUsermodel(\ohist)}(\param{a}\cdot t + \param{b})$, where $\realUsermodel(\ohist)$ is the distribution of (estimated) utility values of the bids in $\ohist$, $Q_{\realUsermodel(B_o(t))}(p)$ is the quantile function of such distribution, and $\param{a}$ and $\param{b}$ are learnable parameters. In other words, we consider the $p^{th}$ best utility received from the agent, where $p$ is a learnable (linear) function of the negotiation time $t$. In this way, this tactic automatically and dynamically decides how much the agent should concede at $t$. Here, $\mathbf{p}_{i,j} = \{a, b\}$ .
\item $\uthresh$, the dynamic DRL-based utility threshold.
\item $u$, a fixed utility threshold.
\end{itemize}

The bidding tactics are:
\begin{itemize}
\item $b_{\mathit{Boulware}}$, a bid generated by a time-dependent Boulware strategy~\cite{fatima2001optimal}.
\item $PS(\param{a}\cdot t + \param{b})$ extracts a bid from the set of Pareto-optimal bids $PS$, derived using the \textit{NSGA-II algorithm}\footnote{Meta-heuristics (instead of brute-force) for Pareto-optimal solutions have the potential to deal efficiently with continuous issues.}~\cite{deb2002fast} under $\realUsermodel$ and $\estOppmodel$.
In particular, it selects the bid that assigns a weight of $\param{a}\cdot t + \param{b}$ to our agent utility (and $1-(\param{a}\cdot t + \param{b})$ to the opponent's), where $\param{a}$ and $\param{b}$ are learnable parameters telling how this weight scales with the negotiation time $t$. The \textit{TOPSIS algorithm}~\cite{hwang1981methods}  is used to derive such a bid, given the weighting $\param{a}\cdot t + \param{b}$ as input. Here, $\param{\mathbf{p}_{i,j} = \{a, b\}}$ .
\item $b_{opp}(\lastobid)$, a tactic to generate a bid by manipulating the last bid received from the opponent $\lastobid$. This is modified in a greedy fashion by randomly changing the value of the least relevant issue (w.r.t.\ ${U}$) of $\lastobid$. 
\item $\omega \sim \mathcal{U}(\Omega_{\geq \uthresh})$, a random bid above $\uthresh$\footnote{$\mathcal{U}(S)$ is the uniform distribution over $S$, and $\Omega_{\geq \uthresh}$ is the subset of $\Omega$ whose bids have estimated utility above $\uthresh$ w.r.t.\ ${U}$.}.
\end{itemize}

For more details, one can refer to~\cite{bagga2021agent, bagga2020learnable}.
We give next an example of a concrete acceptance strategy learned with DLST-ANESIA model for a domain called Party~\footnote{In the Party domain, two agents representing two people while organizing a party negotiates over 6 issues: the food type, drinks type, location, type of invitations, music and the clean up service. Each issue further consists of 3 to 5 values, resulting in a domain with 3072 total possible outcomes.}. 
\begin{small}
\begin{align*}
t \in [0.000, 0.0361) \rightarrow & \
U_u(\omega_t^o) \geq \max\left(Q_{U_{\Omega^o_t}} (-0.20 \cdot t + 0.22), \bar{u_t}\right) \\
t \in [0.0361, 1.000] \rightarrow &  \
U_u(\omega_t^o) \geq \max\left( u, Q_{U_{\Omega^o_t}} (-0.10 \cdot t + 0.64)\right) 
\end{align*} 
\end{small}
Similarly, we learn the following strategy for the Grocery domain~\footnote{In the Grocery domain, two agents representing two people negotiates in a local supermarket who have different tastes. The domain consists of 5 types of products (or issues): bread, fruit, snacks, spreads and vegetables. Each category has further 4 or 5 products, resulting in a medium-sized domain with 1600 possible outcomes.}: 
\begin{small}
\begin{align*}
t \in [0.000, 0.2164) \rightarrow & \
U_u(\omega_t^o) \geq \max\left( U_u(\omega_t), Q_{U_{\Omega^o_t}} (-0.55 \cdot t + 0.05),\bar{u_t} \right) \\
 t \in [0.2164, 0.3379) \rightarrow &  \
U_u(\omega_t^o) \geq \max\left( U_u(\omega_t), Q_{U_{\Omega^o_t}} (-0.60 \cdot t + 1.40)\right) \\
t \in [0.3379, 1.000] \rightarrow &  \
U_u(\omega_t^o) \geq \max\left( Q_{U_{\Omega^o_t}} (-0.22 \cdot t + 0.29), \bar{u_t}\right)
\end{align*} 
\end{small}
Clearly, these templates, containing mathematical expressions and logical conditions, might be decipherable to those with a mathematical background but are often not clear to a wider audience.
\begin{figure*}
    \centering
    \includegraphics[scale=0.32]{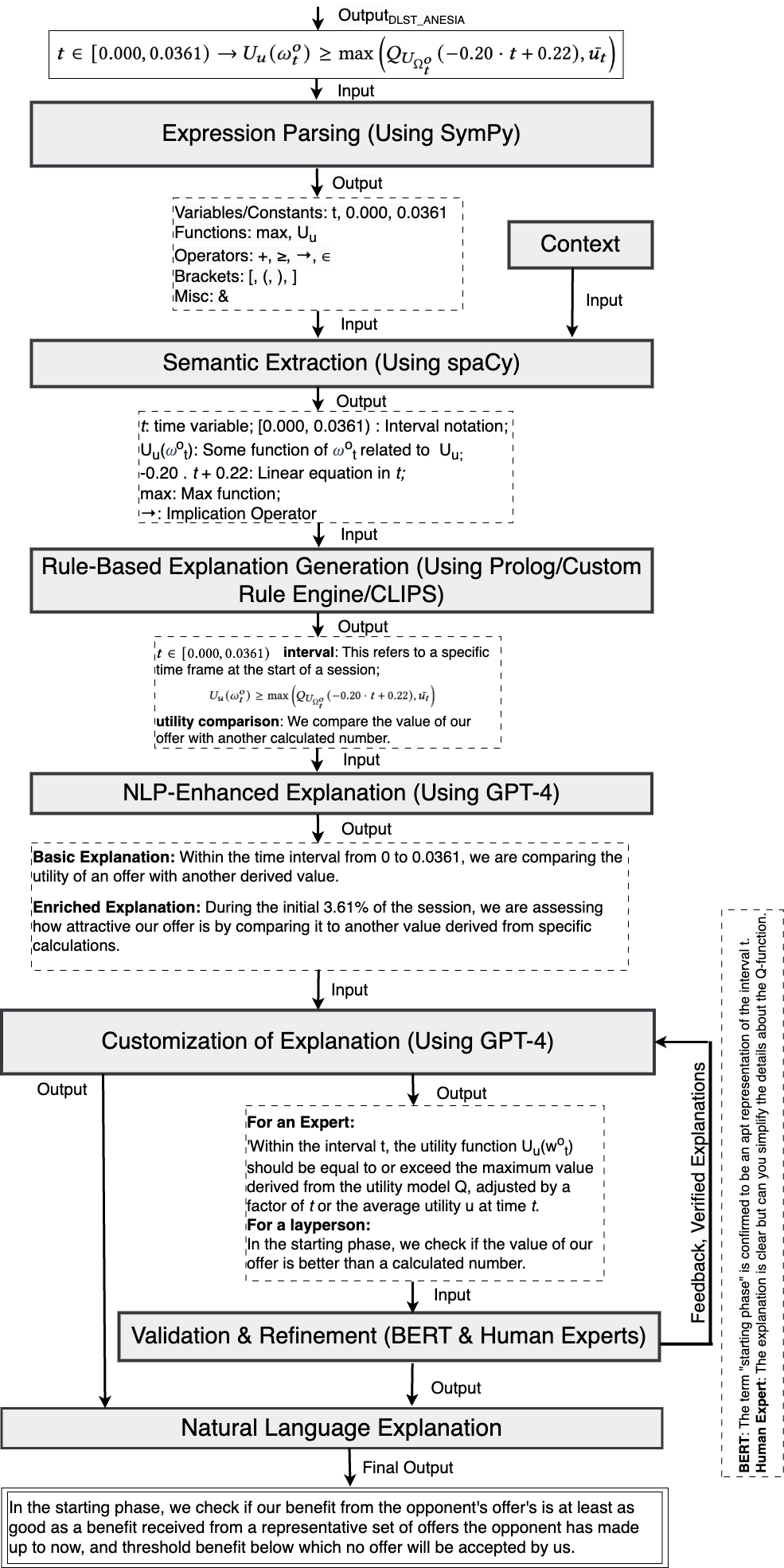}
    \caption{Workflow of the Proposed approach}
    \label{blockdiagram}
\end{figure*}

\section{Explainable Strategy Templates}

The strategy templates generated by our actor-critic DRL framework of the previous section, contain mathematical expressions and logical conditions.
% , which might be interpretable to someone with a mathematical background but may require further explanation to be clear for a wider audience.
Our idea here is to use Natural Language Processing (NLP) to make these expressions explainable and more accessible, by converting them from mathematical expressions and logical rules into clear, plain English sentences.

To automate our idea, we propose a rule-based explainable system that identifies parts of a mathematical expression within a template and maps them to predefined sentence structures in English. 
This can be a quite complex task depending on the variability and complexity of the expressions one would want to handle. Creating a comprehensive, automated system for generating explanations for any given mathematical/logical expression involves a detailed understanding of both the domain (e.g. mathematical expressions/terms specific to say finance) and the model to generate meaningful explanations. As shown in Figure~\ref{blockdiagram}, our proposed approach  encompasses a six-step procedure to convert intricate mathematical expressions into accessible natural language explanations. For clarity, we present a mathematical, algorithmic, and detailed elucidation of each phase.

Let $E$ be a strategy template expression we aim to explain. We will use $P(E)$ for the parsed representation of $E$, $S(P(E))$ for the semantic representation of $P(E)$, $R(S(P(E)))$ for the rule-based sentence structure derived from $S(P(E))$, $T(R(S(P(E))))$ for the enriched explanation using transformers, $C(T(R(S(P(E)))),A)$ for the customized explanation for audience $A$, and $V(C(T(R(S(P(E)))),A))$ be the validated and refined explanation. 
We express this entire process that underlines our approach as\\ 
%$StrategyTemplateToNaturalLanguage(E,A)=V(C(T(R(S(P(E))),A)))$.\\
\begin{equation}
\label{strategy}
    StrategyToNaturalLanguage(E,A)=V(C(T(R(S(P(E))),A)))
\end{equation}

To demonstrate its utility, we will walk through an application of this process using a sample strategy template expression from the Party domain (see Section 2). The idea is to outline the steps, and how they are applied.

\begin{enumerate}[leftmargin=*]

    \item \textbf{Parse the Mathematical Expression} The aim of this step is to  decompose the mathematical expression into identifiable units using Algorithm ~\ref{alg:step1} and our template example. For instance, Variables ($U_u(\omega_t^o)$, $\omega_t^o$, $\Omega^o_t$, $u$, $t$, $\bar{u_t}$), Constants ($-0.20$, $0.22$, $0.000$, $0.361$), Functions (\textit{max}, $U$), Operators ($\geq$, $\in$, $\rightarrow$), and Structure (inequality comparing $U_u(\omega_t^o)$ to the maximum of two expressions. One of the expressions is a function $Q_{U_{\Omega^o_t}}$ of a linear combination of $t$, and the other is $\bar{u_t}$). 
    We employ SymPy~\cite{meurer2017sympy} to parse and symbolically manipulate mathematical expressions.

   \begin{algorithm}
        \caption{Parse Mathematical Expression}\label{alg:step1}
        \begin{algorithmic}[1] % The [1] means every line is numbered
            \Function{ParseMathematicalExpression}{$E$} 
            \State $V = \{\}$,  $C= \{\}$, $F = \{\}$, $O = \{\}$ 
            % \Comment{Set of Variables}
            % \State \Comment{Set of Constants}
            % \State  \Comment{Set of Functions}
            % \State  \Comment{Set of Operators}
            \State $Structure = Tree()$ %\Comment{Representation of hierarchy and relationships}
            \For{each element $e$ in $E$}
                \If{isConstant(e)}  C.add(e)
                \ElsIf{isOperator(e)}  O.add(e)
                \ElsIf{isFunction(e)}  F.add(e)
                \Else\  V.add(e)
                \EndIf
            \EndFor
            \State Structure = ConstructStructure(E, V, C, F, O)
            \State \Return $(V, C, F, O, Structure)$
            \EndFunction
        \end{algorithmic}
    \end{algorithm}

    \item \textbf{Extract Semantic Meaning}
    The goal here is to decode the roles and semantic significance of variables and expression. Using Algorithm~\ref{alg:step2}, we utilize custom logic with NLP (via spaCy~\cite{honnibal2015spacy}) to correlate mathematical entities with semantic roles. In our example template, the variable $t$ can be translated semantically to represent a specific time interval. Specifically, it denotes the initial $3.61\%$ of %an event's 
    a session's duration. This interval provides a crucial context for understanding when the utility of an offer is being assessed.

     \begin{algorithm}
    \caption{Extract Semantic Meaning}\label{alg:step2}
    \begin{algorithmic}[1] % The [1] means every line is numbered
        \Require  Structure, Context
        \Ensure SemanticRep = \{\}
        \Function {ExtractSemanticMeaning}{Structure, Context}
            \For{each node in Structure}
                \State semantic\_tag $\leftarrow$ spaCyIdentifyRole(node) 
                % \Comment{Using spaCy for NLP capabilities}
                \State SemanticRep[node] $\leftarrow$ semantic\_tag
            \EndFor    
        \State \Return SemanticRep
        \EndFunction
    \end{algorithmic}
    \end{algorithm}
    
    \begin{algorithm}
        \caption{Create Rule-Based System}\label{alg:step3}
        \begin{algorithmic}[1]
            \Require SemanticRep
            \Ensure NLTemplate = \{\}
            \Function{CreateRuleBasedSystem}{SemanticRep}
                \State Rules $\leftarrow$ LoadPredefinedRules() 
                % \Comment{Load a predefined set of translation rules}
                \For{each (entity, role) in SemanticRep.items()}
                    \If{role in Rules}
                        \State NLTemplate[entity] $\leftarrow$ Rules[role].apply(entity)
                    \EndIf
                \EndFor
                \State \Return NLTemplate
            \EndFunction
        \end{algorithmic}
    \end{algorithm}
    
    \item \textbf{Create Rule-based System}
    This step establishes different rules for transforming semantic entities and relationships into natural language, as detailed in Algorithm~\ref{alg:step3}. We use a rule-based system translating specific patterns in parsed expressions to predefined linguistic structures (e.g., using Prolog, CLIPS~\cite{clips1985} or a Custom Rule Engine). In the case of our example strategy template, we establish a rule for the \textit{max} function such that it is equivalent to the phrase ``the greater value between X and Y." This means we are seeking the larger of two computed utilities.

    \begin{algorithm}
        \caption{Enrich Explanation}\label{alg:step4}
        \begin{algorithmic}[1]
            \Require NLTemplate
            \Ensure EnrichedExpl = \{\}
            \Function{EnrichExplanation}{NLTemplate}
                \For{each (entity, basicExplain) in NLTemplate.items()}
                    \State enrichedExplain $\leftarrow$ GPT4Elaborate(basicExplain) 
                    % \Comment{Use GPT-4 to elaborate or simplify}
                    \State EnrichedExpl[entity] $\leftarrow$ enrichedExplain
                \EndFor 
               \State \Return EnrichedExpl
            \EndFunction
        \end{algorithmic}
    \end{algorithm}
    
    \item \textbf{Enrich Explanation with Transformers} 
    To enhance our basic rule, we utilize the generative pre-trained transformer GPT-4, with the aim to provide explanations that are rich, nuanced, and akin to human communication. We pass basic explanations derived from Step 3 to GPT-4, requesting elaboration or simplification, as shown in Algorithm~\ref{alg:step4}. 
    The initial phrasing, "Within the time interval of $t$, we are comparing the offer's utility to another derived value," becomes the more detailed ``During the initial $3.61\%$ of the event, we evaluate how the offer's worth compares against a special computed value." 

   \begin{algorithm}
        \caption{Custom Explanation}\label{alg:step5}
        \begin{algorithmic}[1]
            \Require EnrichedExpl, Audience
            \Ensure CustomExpl = \{\}
            \Function{CustomExpl}{EnrichedExpl, Audience}
                \For{each (entity, enrichedExpl) in EnrichedExpl.items()}
                    \If{Audience == `expert'}
                        \State CustomExpl[entity] $\leftarrow$ enrichedExpl
                    \Else
                        \State CustomExpl[entity] $\leftarrow$ SimplifyExpl(enrichedExpl) 
                        % \Comment{Simplification logic}
                    \EndIf
                \EndFor
                
                \Return CustomExpl
            \EndFunction
        \end{algorithmic}
    \end{algorithm}
    
    \item \textbf{Customize Explanation Style}
    Tailoring the explanation to the audience's expertise is vital. As demonstrated in Algorithm~\ref{alg:step5}, given the enriched explanation $T$ and an audience $A$ in~\eqref{strategy}, this step adjusts the explanation to cater to the particular needs, comprehension levels, and terminologies familiar to audience $A$.
     For an expert, we say: "During the interval $t$, the utility function $U_u(\omega_t^o)$ should exceed or equal the computed utility, considering elements like $Q_{U_{\Omega^o_t}}$ and time-based adjustments.'' In contrast, a layperson receives: ``In the beginning phase, we check if our offer's value is at least as good as another number we determine.''
    Another example is:
    For an expert: ``Select the larger value between $u$ and the utility computed from \\ $U_u(\omega_t^o) \geq \max\left( u, Q_{U_{\Omega^o_t}} (-0.10 \cdot t + 0.64)\right)$". For a layperson, the explanation becomes: ``Choose the larger number between `u' and the number we get from our special formula."

     \begin{algorithm}
    \caption{Validate and Refine}\label{alg:step6}
    \begin{algorithmic}[1]
        \Require Explanation
        \Ensure ValidatedExpl = \{\}
        \Function{ValidateAndRefine}{Explanation}
            \For{each (entity, explain) in Explanation.items()}
                \State isValid $\leftarrow$ BERTSemanticValidation(explain, entity) 
                % \Comment{Using BERT for semantic validation}
                \If{isValid}
                    \State ValidatedExpl[entity] $\leftarrow$ explain
                \Else
                    \State ValidatedExpl[entity] $\leftarrow$ RefineExpl(explain) 
                    % \Comment{Refinement logic}
                \EndIf
            \EndFor
            
        \Return ValidatedExplanation
        \EndFunction
    \end{algorithmic}
\end{algorithm}
    
    \item \textbf{Validate and Refine Explanations}
    In the final step outlined in Algorithm~\ref{alg:step6}, we make sure that the generated explanations are validated maintaining their accuracy, clarity and utility. Validated explanations are generated after engaging in a manual review and potentially using Bidirectional Encoder Representations from Transformers (BERT)~\cite{devlin2018bert} for semantic validation, refining explanations and improving system quality. For instance, we must confirm that ``early phase" correctly captures the essence of interval $t$, and that the ``calculated number" genuinely represents the derived utility.

\end{enumerate}

To sum up, our idea of amalgamating the rule-based logic with state-of-the-art transformers (like GPT-4) lends both structure and adaptability to this process.

\section{Conclusions}
We have proposed the amalgamation of NLP techniques and LLMs with transformers like GPT-4 to explore ways of explaining  heuristic strategies for automated negotiation. 
We have employed strategy templates to illustrate the ideas and explored how mathematical expressions of these templates can be translated into a natural language format tailored for application domain users.
% learning mathematical expressions of these templates can then be made accessible in a natural language format for users of the application domain. 
As an example, we have shown the translation of a learned strategy tactic for a negotiation domain from its mathematical form into a coherent narrative, which augments understanding.

In our preliminary exploration of integrating GPT-4 into our workflow, we have found that our method allows for the generation of contextual and human-like explanations from mathematical strategy templates. 
Suitable feedback enables our system to cater for varied audiences and provide an interactive user experience. 
The proposed workflow serves as a foundational step and moves from parsing mathematical expressions through semantic analysis, basic explanation generation enhanced with transformers, interactive query handling, and validation with opportunities of continuous improvement. 
However, ensuring an automated process of comprehensibility and relevance in generated explanations of different mathematical strategies still remains an open challenge.

To address this challenge, our goal is to expand the algorithmic automation of this process, for combining local explanations of tactics to craft a  comprehensive explanation for an entire strategy.

\section*{Acknowledgements} The authors wish to thank the anonymous referees for their comments in a previous version of this paper. The second author was supported by Leverhulme Trust, Research Grant LIP-2022-001.
\iffalse
-- is ther any related work
-- compare to the related work
-- what are the future plans, given this idea
\fi
\bibliographystyle{ACM-Reference-Format} 
\bibliography{references}


\begin{thebibliography}{23}

%%% the Web address.

\ifx \showCODEN    \undefined \def \showCODEN     #1{\unskip}     \fi
\ifx \showDOI      \undefined \def \showDOI       #1{#1}\fi
\ifx \showISBNx    \undefined \def \showISBNx     #1{\unskip}     \fi
\ifx \showISBNxiii \undefined \def \showISBNxiii  #1{\unskip}     \fi
\ifx \showISSN     \undefined \def \showISSN      #1{\unskip}     \fi
\ifx \showLCCN     \undefined \def \showLCCN      #1{\unskip}     \fi
\ifx \shownote     \undefined \def \shownote      #1{#1}          \fi
\ifx \showarticletitle \undefined \def \showarticletitle #1{#1}   \fi
\ifx \showURL      \undefined \def \showURL       {\relax}        \fi
% invisible to TeX
\providecommand\bibfield[2]{#2}
\providecommand\bibinfo[2]{#2}
\providecommand\natexlab[1]{#1}
\providecommand\showeprint[2][]{arXiv:#2}

\bibitem[Alrayes et~al\mbox{.}(2018)]%
        {KAIS17}
\bibfield{author}{\bibinfo{person}{Bedour Alrayes}, \bibinfo{person}{Ozgur Kafali}, {and} \bibinfo{person}{Kostas Stathis}.} \bibinfo{year}{2018}\natexlab{}.
\newblock \showarticletitle{Concurrent {B}ilateral {N}egotiation for {O}pen {E}-{M}arkets: {T}he {CONAN} {S}trategy}.
\newblock \bibinfo{journal}{\emph{Knowledge Information Systems}} \bibinfo{volume}{56}, \bibinfo{number}{2} (\bibinfo{year}{2018}), \bibinfo{pages}{463--501}.
\newblock


\bibitem[Baarslag et~al\mbox{.}(2014)]%
        {baarslag2014decoupling}
\bibfield{author}{\bibinfo{person}{Tim Baarslag}, \bibinfo{person}{Koen Hindriks}, \bibinfo{person}{Mark Hendrikx}, \bibinfo{person}{Alexander Dirkzwager}, {and} \bibinfo{person}{Catholijn Jonker}.} \bibinfo{year}{2014}\natexlab{}.
\newblock \showarticletitle{Decoupling negotiating agents to explore the space of negotiation strategies}.
\newblock In \bibinfo{booktitle}{\emph{Novel Insights in Agent-based Complex Automated Negotiation}}. \bibinfo{publisher}{Springer}, \bibinfo{pages}{61--83}.
\newblock


\bibitem[Bagga(2021)]%
        {bagga2021agent}
\bibfield{author}{\bibinfo{person}{Pallavi Bagga}.} \bibinfo{year}{2021}\natexlab{}.
\newblock \emph{\bibinfo{title}{Agent Learning for Automated Bilateral Negotiations}}.
\newblock \bibinfo{thesistype}{Ph.\,D. Dissertation}. \bibinfo{school}{Royal Holloway, University of London}.
\newblock


\bibitem[Bagga et~al\mbox{.}(2021a)]%
        {anegma}
\bibfield{author}{\bibinfo{person}{Pallavi Bagga}, \bibinfo{person}{Nicola Paoletti}, \bibinfo{person}{Bedour Alrayes}, {and} \bibinfo{person}{Kostas Stathis}.} \bibinfo{year}{2021}\natexlab{a}.
\newblock \showarticletitle{{ANEGMA}: an automated negotiation model for e-markets}.
\newblock \bibinfo{journal}{\emph{Journal of Autonomous Agents and Multi-Agent Systems}}  \bibinfo{volume}{35} (\bibinfo{year}{2021}).
\newblock


\bibitem[Bagga et~al\mbox{.}(2020)]%
        {bagga2020learnable}
\bibfield{author}{\bibinfo{person}{Pallavi Bagga}, \bibinfo{person}{Nicola Paoletti}, {and} \bibinfo{person}{Kostas Stathis}.} \bibinfo{year}{2020}\natexlab{}.
\newblock \showarticletitle{Learnable strategies for bilateral agent negotiation over multiple issues}.
\newblock \bibinfo{journal}{\emph{arXiv preprint arXiv:2009.08302}} (\bibinfo{year}{2020}).
\newblock


\bibitem[Bagga et~al\mbox{.}(2021b)]%
        {bagga2021pareto}
\bibfield{author}{\bibinfo{person}{Pallavi Bagga}, \bibinfo{person}{Nicola Paoletti}, {and} \bibinfo{person}{Kostas Stathis}.} \bibinfo{year}{2021}\natexlab{b}.
\newblock \showarticletitle{Pareto Bid Estimation for Multi-Issue Bilateral Negotiation under User Preference Uncertainty}. In \bibinfo{booktitle}{\emph{2021 IEEE International Conference on Fuzzy Systems (FUZZ-IEEE)}}. IEEE, \bibinfo{pages}{1--6}.
\newblock


\bibitem[Bagga et~al\mbox{.}(2022)]%
        {bagga2022deep}
\bibfield{author}{\bibinfo{person}{Pallavi Bagga}, \bibinfo{person}{Nicola Paoletti}, {and} \bibinfo{person}{Kostas Stathis}.} \bibinfo{year}{2022}\natexlab{}.
\newblock \showarticletitle{Deep learnable strategy templates for multi-issue bilateral negotiation}. In \bibinfo{booktitle}{\emph{Proc. of the 21st International Conference on Autonomous Agents and Multiagent Systems (AAMAS 2022)}}, \bibfield{editor}{\bibinfo{person}{P.~Faliszewski}, \bibinfo{person}{V.~Mascardi}, \bibinfo{person}{C.~Pelachaud}, {and} \bibinfo{person}{M.E. Taylor}} (Eds.).
\newblock


\bibitem[Brown et~al\mbox{.}(2020)]%
        {brown2020language}
\bibfield{author}{\bibinfo{person}{Tom~B Brown}, \bibinfo{person}{Benjamin Mann}, {et~al\mbox{.}}} \bibinfo{year}{2020}\natexlab{}.
\newblock \showarticletitle{Language models are few-shot learners}.
\newblock \bibinfo{journal}{\emph{arXiv preprint arXiv:2005.14165}} (\bibinfo{year}{2020}).
\newblock


\bibitem[Deb et~al\mbox{.}(2002)]%
        {deb2002fast}
\bibfield{author}{\bibinfo{person}{Kalyanmoy Deb}, \bibinfo{person}{Amrit Pratap}, \bibinfo{person}{Sameer Agarwal}, {and} \bibinfo{person}{TAMT Meyarivan}.} \bibinfo{year}{2002}\natexlab{}.
\newblock \showarticletitle{A fast and elitist multiobjective genetic algorithm: NSGA-II}.
\newblock \bibinfo{journal}{\emph{IEEE transactions on evolutionary computation}} \bibinfo{volume}{6}, \bibinfo{number}{2} (\bibinfo{year}{2002}), \bibinfo{pages}{182--197}.
\newblock


\bibitem[Devlin et~al\mbox{.}(2018)]%
        {devlin2018bert}
\bibfield{author}{\bibinfo{person}{Jacob Devlin}, \bibinfo{person}{Ming-Wei Chang}, \bibinfo{person}{Kenton Lee}, {and} \bibinfo{person}{Kristina Toutanova}.} \bibinfo{year}{2018}\natexlab{}.
\newblock \showarticletitle{BERT: Pre-training of Deep Bidirectional Transformers for Language Understanding}.
\newblock \bibinfo{journal}{\emph{arXiv preprint arXiv:1810.04805}} (\bibinfo{year}{2018}).
\newblock


\bibitem[Fatima et~al\mbox{.}(2001)]%
        {fatima2001optimal}
\bibfield{author}{\bibinfo{person}{S~Shaheen Fatima}, \bibinfo{person}{Michael Wooldridge}, {and} \bibinfo{person}{Nicholas~R Jennings}.} \bibinfo{year}{2001}\natexlab{}.
\newblock \showarticletitle{Optimal negotiation strategies for agents with incomplete information}. In \bibinfo{booktitle}{\emph{{ATAL}'01}}. Springer, \bibinfo{pages}{377--392}.
\newblock


\bibitem[Fatima et~al\mbox{.}(2005)]%
        {fatima2005comparative}
\bibfield{author}{\bibinfo{person}{Shaheen~S Fatima}, \bibinfo{person}{Michael Wooldridge}, {and} \bibinfo{person}{Nicholas~R Jennings}.} \bibinfo{year}{2005}\natexlab{}.
\newblock \showarticletitle{A comparative study of game theoretic and evolutionary models of bargaining for software agents}.
\newblock \bibinfo{journal}{\emph{Artificial Intelligence Review}} \bibinfo{volume}{23}, \bibinfo{number}{2} (\bibinfo{year}{2005}), \bibinfo{pages}{187--205}.
\newblock


\bibitem[Honnibal and Montani(2015)]%
        {honnibal2015spacy}
\bibfield{author}{\bibinfo{person}{Matthew Honnibal} {and} \bibinfo{person}{Ines Montani}.} \bibinfo{year}{2015}\natexlab{}.
\newblock \bibinfo{title}{spaCy: Industrial-strength Natural Language Processing in Python}.
\newblock \bibinfo{howpublished}{\url{https://spacy.io}}.
\newblock


\bibitem[Hwang and Yoon(1981)]%
        {hwang1981methods}
\bibfield{author}{\bibinfo{person}{Ching-Lai Hwang} {and} \bibinfo{person}{Kwangsun Yoon}.} \bibinfo{year}{1981}\natexlab{}.
\newblock \showarticletitle{Methods for multiple attribute decision making}.
\newblock In \bibinfo{booktitle}{\emph{Multiple attribute decision making}}. \bibinfo{publisher}{Springer}, \bibinfo{pages}{58--191}.
\newblock


\bibitem[Liu et~al\mbox{.}(2021)]%
        {liu2021finrl}
\bibfield{author}{\bibinfo{person}{Xiao-Yang Liu}, \bibinfo{person}{Hongyang Yang}, \bibinfo{person}{Jiechao Gao}, {and} \bibinfo{person}{Christina~Dan Wang}.} \bibinfo{year}{2021}\natexlab{}.
\newblock \showarticletitle{FinRL: Deep reinforcement learning framework to automate trading in quantitative finance}. In \bibinfo{booktitle}{\emph{Proceedings of the second ACM international conference on AI in finance}}. \bibinfo{pages}{1--9}.
\newblock


\bibitem[Liu et~al\mbox{.}(2019)]%
        {liu2019roberta}
\bibfield{author}{\bibinfo{person}{Yinhan Liu}, \bibinfo{person}{Myle Ott}, {et~al\mbox{.}}} \bibinfo{year}{2019}\natexlab{}.
\newblock \showarticletitle{Roberta: A robustly optimized bert pretraining approach}.
\newblock \bibinfo{journal}{\emph{arXiv preprint arXiv:1907.11692}} (\bibinfo{year}{2019}).
\newblock


\bibitem[Meurer et~al\mbox{.}(2017)]%
        {meurer2017sympy}
\bibfield{author}{\bibinfo{person}{Aaron Meurer}, \bibinfo{person}{Christopher~P Smith}, {et~al\mbox{.}}} \bibinfo{year}{2017}\natexlab{}.
\newblock \showarticletitle{SymPy: symbolic computing in Python}.
\newblock \bibinfo{journal}{\emph{PeerJ Computer Science}}  \bibinfo{volume}{3} (\bibinfo{year}{2017}), \bibinfo{pages}{e103}.
\newblock
\urldef\tempurl%
\url{https://www.sympy.org}
\showURL{%
\tempurl}


\bibitem[{NASA Software}(1985)]%
        {clips1985}
\bibfield{author}{\bibinfo{person}{{NASA Software}}.} \bibinfo{year}{1985}\natexlab{}.
\newblock \bibinfo{booktitle}{\emph{C Language Integrated Production System (CLIPS)}}.
\newblock NASA Lyndon B. Johnson Space Center, Houston, Texas.
\newblock
\urldef\tempurl%
\url{https://www.clipsrules.net/}
\showURL{%
\tempurl}
\newblock
\shownote{Version 6.31}.


\bibitem[OpenAI(2023)]%
        {openai2023gpt4}
\bibfield{author}{\bibinfo{person}{OpenAI}.} \bibinfo{year}{2023}\natexlab{}.
\newblock \showarticletitle{GPT-4: Technical Report}.
\newblock \bibinfo{journal}{\emph{arXiv preprint arXiv:4812508}} (\bibinfo{year}{2023}).
\newblock
\urldef\tempurl%
\url{https://cdn.openai.com/papers/gpt-4.pdf}
\showURL{%
\tempurl}


\bibitem[Raffel et~al\mbox{.}(2020)]%
        {raffel2020exploring}
\bibfield{author}{\bibinfo{person}{Colin Raffel}, \bibinfo{person}{Noam Shazeer}, {et~al\mbox{.}}} \bibinfo{year}{2020}\natexlab{}.
\newblock \showarticletitle{Exploring the limits of transfer learning with a unified text-to-text transformer}.
\newblock \bibinfo{journal}{\emph{The Journal of Machine Learning Research}} \bibinfo{volume}{21}, \bibinfo{number}{1} (\bibinfo{year}{2020}), \bibinfo{pages}{5485--5551}.
\newblock


\bibitem[Rubinstein(1982)]%
        {rubinstein1982perfect}
\bibfield{author}{\bibinfo{person}{Ariel Rubinstein}.} \bibinfo{year}{1982}\natexlab{}.
\newblock \showarticletitle{Perfect equilibrium in a bargaining model}.
\newblock \bibinfo{journal}{\emph{Econometrica: Journal of the Econometric Society}} (\bibinfo{year}{1982}), \bibinfo{pages}{97--109}.
\newblock


\bibitem[Van Der~Putten et~al\mbox{.}(2006)]%
        {van2006automating}
\bibfield{author}{\bibinfo{person}{Sander Van Der~Putten}, \bibinfo{person}{Valentin Robu}, \bibinfo{person}{Han La~Poutr{\'e}}, \bibinfo{person}{Annemiek Jorritsma}, {and} \bibinfo{person}{Margo Gal}.} \bibinfo{year}{2006}\natexlab{}.
\newblock \showarticletitle{Automating supply chain negotiations using autonomous agents: a case study in transportation logistics}. In \bibinfo{booktitle}{\emph{Proceedings of the fifth international joint conference on Autonomous agents and multiagent systems}}. \bibinfo{pages}{1506--1513}.
\newblock


\bibitem[Yang et~al\mbox{.}(2019)]%
        {yang2019xlnet}
\bibfield{author}{\bibinfo{person}{Zhilin Yang}, \bibinfo{person}{Zihang Dai}, {et~al\mbox{.}}} \bibinfo{year}{2019}\natexlab{}.
\newblock \showarticletitle{Xlnet: Generalized autoregressive pretraining for language understanding}.
\newblock \bibinfo{journal}{\emph{Advances in neural information processing systems}}  \bibinfo{volume}{32} (\bibinfo{year}{2019}).
\newblock


\end{thebibliography}

\end{document}